%% file: main.tex
\documentclass{article}

\usepackage{resources/PRIMEarxiv}

\usepackage[utf8]{inputenc} % allow utf-8 input
\usepackage[T1]{fontenc}    % use 8-bit T1 fonts
\usepackage{hyperref}       % hyperlinks
\usepackage{url}            % simple URL typesetting
\usepackage{booktabs}       % professional-quality tables
\usepackage{amsfonts}       % blackboard math symbols
\usepackage{nicefrac}       % compact symbols for 1/2, etc.
\usepackage{microtype}      % microtypography
\usepackage{lipsum}
\usepackage{fancyhdr}       % header
\usepackage{graphicx}       % graphics
\usepackage{multirow}
\usepackage{cleveref}
\usepackage{color,soul}
\usepackage{wrapfig}
\graphicspath{{media/}}     % organize your images and other figures under media/ folder

\title{Evaluation Methodology for Large Language Models for Multilingual Document Question and Answer
 \thanks{\textit{\underline{Correspondence}}: 
 \textbf{ {adarkahana, jaymathe}@microsoft.com}} 
}

\author{
  Adar Kahana, Jaya Susan Mathew, Said Bleik, Jeremy Reynolds, Oren Elisha \\
  Microsoft Corporation \\
  \texttt{\{adarkahana, jaymathe, bleik, jeremr, orelisha\}@microsoft.com} \\
   \And
}

\begin{document}
\maketitle

\begin{abstract}
With the widespread adoption of Large Language Models (LLMs), in this paper we investigate the multilingual capability of these models. Our preliminary results show that, translating the native language context, question and answer into a high resource language produced the best results. 
\end{abstract}

% keywords can be removed
\keywords{Large Language Models (LLM) \and Generative Pretrained Transformers (GPT) \and ChatGPT \and Multilingual support \and Multilingual model evaluation}

\input{introduction}

\input{methodology}

\input{results}

\section{Conclusion}

We proposed a useful method for evaluating the performance of LLMs in multilingual setups. We presented results for several scenarios that are based on the evaluation processes proposed for testing internal models before publishing them as various components of customer-facing products. A summary of augmentations that surfaced from this study is as follows:

\begin{itemize}
    \item In multilingual scenarios, it is preferable to operate in English (if possible). This introduces extra cost, either by extra calls to LLMs for translation or by using a translation service, but improves the results across the board.
    \item There is a large gap between the various GPT versions. Using the latest models are justified in a multilingual scenario.
    \item Datasets that are naturally in a different language are much harder for the given task, but GPT-4 gives decent results, especially when operating in English (including translations).
\end{itemize}

\section*{Acknowledgments}

This work has been performed by members of the Applied Science team at Microsoft Industry AI.

%Bibliography
\bibliographystyle{unsrt}  
\bibliography{main}

\end{document}

%% file: introduction.tex
\section{Introduction}

With the publication of the paper, `Attention is All You Need' \cite{vaswani2017attention}, transformer architecture and attention mechanism has made way for a plethora of Large Language Models (LLMs). More recently with the launch of ChatGPT (Chat Generative Pre-trained Transformer) \cite{ChatGPT}, there has been a growing interest amongst the general public as well as in large businesses in using these LLMs in improving their efficiency \cite{bahrini2023chatgpt} in various common scenarios like summarizing a document, answering a question, solving a mathematics problem to even writing code. 

\begin{wrapfigure}{r}{0.35\textwidth}
    \centering
    \includegraphics[width=0.35\textwidth]{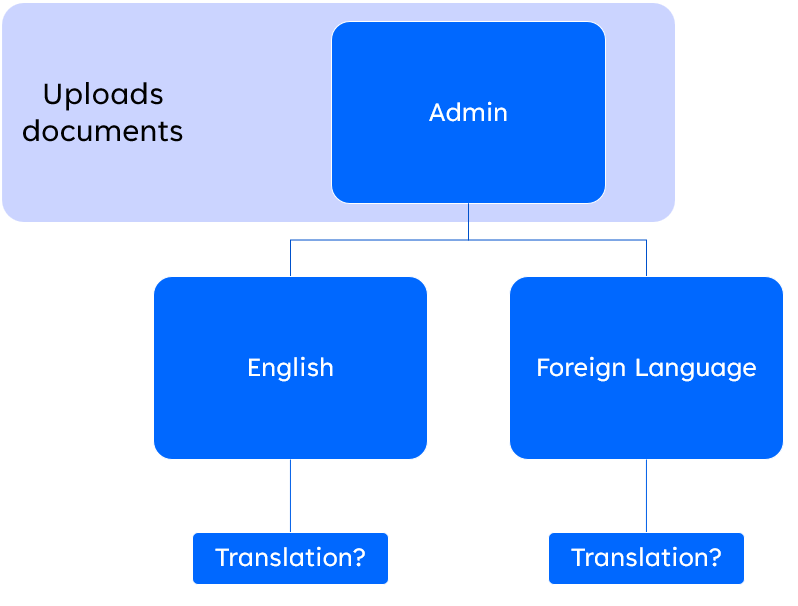}
    \caption{Admin uploading files for Question-Answering module that can be translated either to or from English}
    \label{fig:admin_qna}
\end{wrapfigure}

Majority of these LLMs are pre-trained using predominantly datasets in English and some high resource languages \cite{yuan2023multilingual} \cite{wiegreffe2021teach}, hence tend to perform best in English and in these high resource languages but tend to degrade in their performance in other especially low resource languages like some of the languages spoken in Asia and Africa \cite{zhu2023multilingual}. However, these high resource languages do not necessarily account for majority of the global population. To enable widespread adoption of these LLMs around the world we would need to ensure that these models can support multiple languages in addition to the population who understand and can converse in English or these high resource languages \cite{julian2020most} \cite{mostspokenlanguages}. In addition, businesses and organizations are looking to using these models on a global scale to cater to their consumers around their world in the language of their choice \cite{Worldlanguages}. 

To address this issue and enhance language support for these LLMs, there is ongoing research on whether the underlying model needs to be trained from scratch using multilingual data or whether fine-tuning an existing model with sample multilingual data will suffice or whether some simple effective prompt engineering techniques will be sufficient or whether we need to translate documents into a high resource language to enable multilingual support \cite{shi2022language} \cite{virtanen2019multilingual} \cite{rust2020good} \cite{ebrahimi2021adapt} \cite{pires2019multilingual} \cite{kakwani2020indicnlpsuite} \cite{tang2020multilingual}. There are parallel ongoing efforts to collect and label data in multiple languages including the low resource languages to improve the training corpus. Evaluating multilingual model performance is also an area of active research since most of the popular model performance benchmarks are also predominately for the English language \cite{srinivasan2021predicting} \cite{hu2020xtreme}. 

\begin{wrapfigure}{l}{0.5\textwidth}
    \centering
    \includegraphics[width=0.5\textwidth]{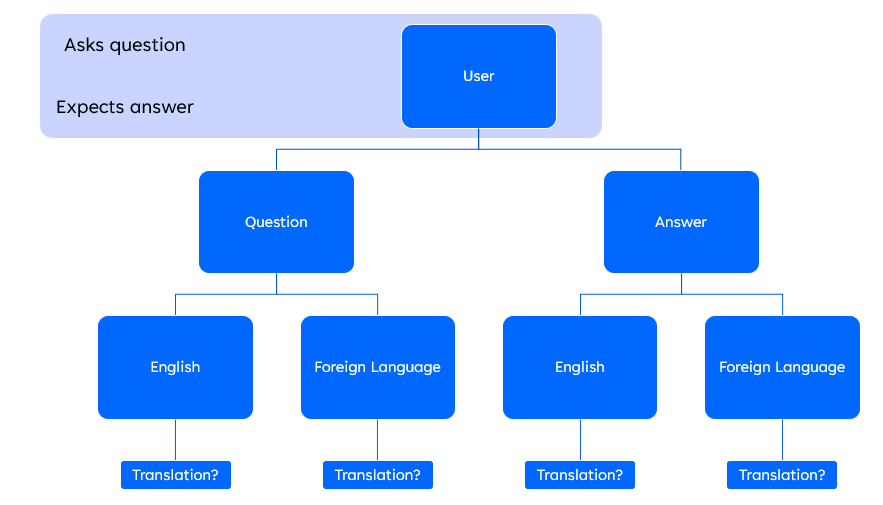}
    \caption{Admin uploading files for Question-Answering module that can be translated either to or from English}
    \label{fig:user_qna}
\end{wrapfigure}

In this paper, we evaluate the multilingual support for select Generative Pre-trained Transformer (GPT) models on the Question-Answering task using multiple datasets like the Stanford Question Answering Dataset (SQuAD) \cite{rajpurkar2016squad}, Cross-lingual Question Answering Dataset (XQuAD) \cite{artetxe2019cross}, Environmental, Social and Governance sustainability Dataset (ESG) and the Hebrew Question Answering Dataset (HeQ) \cite{cohen2023heq}. These datasets enables us to test our GPT model performance and evaluate their performance across multiple languages. The Question-Answering task of interest can be broken into two, with a real application as an example: an admin uploads a set of documents for customers to query and get answers based on these documents. This flow has several possible points where translation can be applied. An illustration of this process is given in \cref{fig:admin_qna} and \cref{fig:user_qna}. We aim to supply informative insights on the quality of these models in a multilingual Question-Answering scenario, and investigate the effects of translation on several key components of the illustrated process. We conclude with recommendations based on the evaluations we made on the available datasets.

%% file: methodology.tex
\section{Methodology}

\subsection{Evaluation flow}

We propose a methodology for testing the quality of Question-Answering tasks using different LLMs. In this work, we present results when using GPT-4-32K \cite{gpt4models} and GPT-3.5-Turbo \cite{gpt3.5models} models. We use the Azure OpenAI service \cite{azureopenai} and the tests were run on a Python environment, where we control the entirety of the process except for the chat completion calls made to the LLMs. 

The flow of the tests starts with randomly selecting a subset from these datasets, where each sample has a context, a question, and one or more answers. The context is either a small paragraph that if given to a human, they would read it and be able to answer the question, or a document that gives sufficient information to a human attempting to answer the question. The second part of the flow involves querying the LLM to answer the question. We use simple prompts like `Here is context for the question: ' and `Please, given the context, answer the following question:', followed by the context and question respectively. We emphasize that even though the system prompts are in English, the contexts and questions are injected in any foreign language.

Lastly, we ask the LLM to verify if the generated answer is correct. To do this, we use a system prompt like `Is your answer correct? The `true' and correct answer is: 
X. Your answer is: Y. Reply Only `TRUE' IF `YES' OR `FALSE' IF `NO''. We concatenate the `true' answer from the dataset in the place of X and the inferred answer by the LLM in the place of Y. Then we look for the word `True' (checking for upper/lower case as well), as well as `Yes' which interestingly sometimes gets returned instead of `True' (and `No' instead of `False'). In the case of multiple answers, we iterate over all the answers and if a correct answer has been supplied by the LLM, we advance the counter and continue to the next question.

\subsection{Datasets}

We explored the following datasets:

\begin{itemize}
    \item \textbf{XQuAD}: Cross-lingual Question Answering Dataset (XQuAD) \cite{artetxe2019cross}. This dataset includes 12 languages and 1,190 context-question-answers samples. The topics are generic and vary between many fields of interest. An important aspect of this dataset is that each question has been translated from English to all these languages (Arabic, German, Greek, Spanish, Hindi, Russian, Thai, Turkish, Vietnamese, Chinese, Romanian) with a high level of confidence in the translation, as it is translated by human translators. We randomly select 50 question from this dataset to run the experiments, ensuring that they are the same questions across the different languages.
    \item \textbf{SQuAD}: The Stanford Question Answering Dataset (SQuAD) \cite{rajpurkar2016squad}. This dataset is part of the GLUE benchmark \cite{wang2018glue} \cite{gluebenchmark}, which has been a standard of testing for many models, and even part of the training dataset of some of the popular LLMs. The dataset is completely in English. With this dataset we explore the translation capabilities of these LLMs. In the evaluation flow, we add a step to translate the context, the question and the true answers, but using the system prompt `Please translate the following questions to LANGUAGE: ', as an example for translation of the questions to language LANGUAGE. The list of questions in this case is concatenated to the end of the system prompt, resulting in a list of translated questions. This dataset has 98,169 context-question-answers samples and we select 50 random questions for the experiments.
    \item \textbf{ESG}: Environmental, Social and Governance sustainability Dataset (ESG). This dataset includes nine documents, in PDF format, which are annual ESG reports of different corporations like Microsoft \cite{MicrosoftESG}, Amazon \cite{AmazonESG}, etc. We have access to 56 questions and answers related to these documents, that are internal Microsoft data. We report the findings on this real world industrial dataset.
    \item \textbf{HeQ}: Hebrew Question Answering Dataset. This dataset follows the format and crowd-sourcing methodology of the SQuAD and the original ParaShoot Datasets \cite{keren2021parashoot}. A team of crowd-workers formulated and answered reading comprehension questions based on random paragraphs in Hebrew \cite{cohen2023heq}. The paragraphs are sourced from two different platforms: (1) Hebrew Wikipedia, and (2) Geektime, an online Israeli news channel specializing in technology. Two types of questions were collected namely: `Answerable' questions (21K) and `Unanswerable' questions (8K) wherein the `Answerable' questions had answers present in the paragraph while the `Unanswerable' questions did not have the answers explicitly included in the paragraph.
\end{itemize}

\subsection{Experiments}

We carried out three different experiments. The experiments differ in the origins of the dataset, which surface the needs for translation from or to English. The first experiment involves the XQuAD dataset, where the context-question-answers samples are originally in English and about the North American culture, industry, etc. The translated versions include translated contexts, as well as the questions and true answers, which means we do not need to translate anything in the process. This sets the benchmark for the performance. In addition, we run the English tests 10 times and compute the mean accuracy and the standard deviation across all 10 tests, to ensure that the proposed testing framework is consistent. The results show a small percentage (up to $~4\%$) of standard deviation, which ensures consistency.

The second experiment involves using the SQuAD dataset, which is an English only dataset but similar in nature to the XQuAD dataset. In this case we want to experiment with machine translation, so we inject translation pieces into the pipeline as discussed in the proposed evaluation methodology. In this experiment we conduct another test, aimed at isolating and investigating the translation pieces. This test involves using the English contexts and questions, receiving a question in English and translating it to the foreign language for evaluation. It mimics, for example, a scenario where the data is in English (e.g., fetched from a web based English corpus) but the answer returned to the user should be in their foreign language. We refer to it as `partial translation' in the results section.

The third experiment involves the ESG dataset, and the pipeline is rather similar to the second experiment. However, it is worth noting that the dataset has PDF documents as contexts (in contrast to clean string paragraphs). We use a `PDF to text' converter, which results in lots of long and noisy texts. As a fair comparison, we use the text files for the English version, as well as translate them using the translation functionality of the GPT model, to obtain and use text files in other languages as well. The translated text files are cleaner than the English ones, but may suffer from early-stopping of the GPT, chopping the majority of the document. We report the findings after chopping, as this is an important finding by itself that brings insights to scientist who use these models for translation as part of a pipeline.

The last experiment involves the HeQ dataset, which is originally in Hebrew and discusses various aspects of the Israeli culture, industry, etc. It is similar in structure to XQuAD and SQuAD, but the content makes it challenging. In this case, a translation means from Hebrew to English. To make a fair comparison, we substitute only the translated contents but keep the same prompts, making it susceptible to `language barrier' issues should the models have any. Example issues are correct answers but failure to recognize it as the words change meaning with pronunciation, connecting words that change the meaning of the word, and more. We observe those when selecting several failing samples, and decide to include them when calculating the overall accuracy.

%% file: results.tex
\section{Results}

We carried out the first experiment and gathered the results in \cref{table:exp_1}. We mention that to run each language takes on average of approximately 20 minutes with GPT-4 and about five with GPT-3.5. We observe that there is a major leap in accuracy when using GPT-4. In addition, from the English test we see that the standard deviation of the error is rather small, and the results are very consistent. We also observe that in most cases both GPT-4 and GPT-3.5 either struggled with the same language, such as in Greek and Hindi, or excelled with the language, such as in Spanish.

\begin{table}[!ht]
    \centering
    \begin{tabular}{ |p{3cm}||p{5cm}|p{5cm}|  }
        \hline
        \textbf{Language} & \textbf{Accuracy using GPT-4-32K} & \textbf{Accuracy using GPT-3.5-Turbo} \\
        \hline
        English & $85.6\% \pm 3.8\%$ & $65.6\% \pm 4.7\%$ \\
        Spanish & $84\%$ & $62\%$ \\
        German & $74\%$ & $44\%$ \\
        Greek & $68\%$ & $24\%$ \\
        Russian & $76\%$ & $36\%$ \\
        Turkish & $68\%$ & $40\%$ \\
        Arabic & $70\%$ & $22\%$ \\
        Vietnamese & $82\%$ & $46\%$ \\
        Thai & $62\%$ & $28\%$ \\
        Chinese & $72\%$ & $52\%$ \\
        Hindi & $68\%$ & $22\%$ \\
        \hline
    \end{tabular}
    \caption{Localization results, XQuAD dataset}
    \label{table:exp_1}
\end{table}

The results of the second experiment are given in \cref{table:exp_2}. We observe clearly for GPT-4 that partial translation works better than full translation, which advocated for the use of English. This provides more evidence to the hypothesis that each translation piece adds complications to the pipeline. We also observe poor performance of GPT-3.5 in this case, which can be explained by the additional calls to the model (extra translation components done using GPT-3.5 as well), each has it's own error, and the errors tend to accumulate. As expected, the English results conform with those of the XQuAD.

\begin{table}[!ht]
    \centering
    \begin{tabular}{ |p{5cm}||p{5cm}|p{5cm}|  }
        \hline
        \textbf{Language} & \textbf{Accuracy using GPT-4-32K} & \textbf{Accuracy using GPT-3.5-Turbo} \\
        \hline
        English & $85.3\% \pm 5.2\%$ & $28.7\% \pm 13.1\%$ \\
        Dutch (Full translation) & $74\%$ & $32\%$ \\
        Dutch (Partial translation) & $78\%$ & $22\%$ \\
        German (Full translation) & $72\%$ & $44\%$ \\
        German (Partial translation) & $80\%$ & $40\%$ \\
        Hebrew (Full translation) & $60\%$ & $28\%$ \\
        Hebrew (Partial translation) & $60\%$ & $38\%$ \\
        \hline
    \end{tabular}
    \caption{Localization results, SQuAD dataset with translation}
    \label{table:exp_2}
\end{table}

For the third experiment, and since it is on real data, we observe lower performance as expected, compared to the former cleaner experiments. The accumulated error is a largely contributing factor for the drop in performance. We do observe, however, that GPT-4 is still able to perform on this dataset, while the GPT-3.5 gets almost all questions wrong. We recall that the translation of the documents also `cleans' the text (removing unwanted characters), but some of the documents in the foreign language have been chopped by the LLM. The results are given in \cref{table:exp_3}

\begin{table}[!ht]
    \centering
    \begin{tabular}{ |p{5cm}||p{5cm}|p{5cm}|  }
        \hline
        \textbf{Language} & \textbf{Accuracy using GPT-4-32K} & \textbf{Accuracy using GPT-3.5-Turbo} \\
        \hline
        English & $63.4\% \pm 2.7\%$ & $18.75\% \pm 4.5\%$ \\
        Dutch (Full translation) & $75\%$ & $28.6\%$ \\
        Dutch (Partial translation) & $60.7\%$ & $12.5\%$ \\
        German (Full translation) & $58.9\%$ & $32.14\%$ \\
        German (Partial translation) & $53.57\%$ & $12.5\%$ \\
        \hline
    \end{tabular}
    \caption{Localization results, ESG dataset with translation}
    \label{table:exp_3}
\end{table}

The last experiment shows very interesting results, shown in \cref{table:exp_4}. The dataset is in Hebrew and related to Israel and Hebrew questions and contexts, is very less likely to have been part of the training of the LLMs. However, despite the poor performance of the GPT-3.5 model, the GPT-4 model shows decent performance. In addition, another interesting insight is that translating the data into English and performing the question answering in English yields even better results. We observed a similar phenomenon with a small Hindi test dataset as well but did not include a thorough explanation for this dataset as it is a small one with only five context-question-answer samples, manually created for a specific document Question-Answering task. The same phenomenon is clearly observed for all the five questions in Hindi, which encourages this recommendation.

\begin{table}[!ht]
    \centering
    \begin{tabular}{ |p{5cm}||p{5cm}|p{5cm}|  }
        \hline
        \textbf{Language} & \textbf{Accuracy using GPT-4-32K} & \textbf{Accuracy using GPT-3.5-Turbo} \\
        \hline
        Hebrew & $56\%$ & $24\%$ \\
        English (Full translation) & $64\%$ & $28\%$ \\
        English (Partial translation) & $58\%$ & $18\%$ \\
        \hline
    \end{tabular}
    \caption{Localization results, HeQ dataset with translation}
    \label{table:exp_4}
\end{table}